\documentclass{article}

\usepackage{microtype}
\usepackage{graphicx}
\usepackage{subcaption}
\usepackage{booktabs} 
\usepackage{booktabs}   
\usepackage[table]{xcolor} 
\usepackage{array}      
\usepackage{multirow}
\usepackage{tcolorbox}
\usepackage{hyperref}

\definecolor{rowgray}{gray}{0.9}

\usepackage[preprint]{icml2026}

\usepackage{amsmath}
\usepackage{amssymb}
\usepackage{mathtools}
\usepackage{amsthm}

\usepackage[capitalize,noabbrev]{cleveref}

\theoremstyle{plain}

\theoremstyle{definition}

\theoremstyle{remark}

\usepackage[textsize=tiny]{todonotes}

\icmltitlerunning{ProMAS: Proactive Error Forecasting for Multi-Agent Systems Using Markov Transition Dynamics}

\newcommand{\modelname}{\textsc{ProMAS}}

\begin{document}

\twocolumn[
  \icmltitle{ProMAS: Proactive Error Forecasting for Multi-Agent Systems \\Using Markov Transition Dynamics}



  \icmlsetsymbol{equal}{*}

  \begin{icmlauthorlist}
    \icmlauthor{Xinkui Zhao}{yyy}
    \icmlauthor{Sai Liu}{yyy}
    \icmlauthor{Yifan Zhang}{yyy}
    \icmlauthor{Qingyu Ma}{yyy}
    \icmlauthor{Guanjie Cheng}{yyy}
    \icmlauthor{Naibo Wang}{yyy}
    \icmlauthor{Chang Liu}{yyy}
  \end{icmlauthorlist}

  \icmlaffiliation{yyy}{Zhejiang University}

  \icmlcorrespondingauthor{Yifan Zhang}{12451018@zju.edu.cn}

  \icmlkeywords{Machine Learning, ICML}

  \vskip 0.3in
]



\printAffiliationsAndNotice{}  

\begin{abstract}
  The integration of Large Language Models into Multi-Agent Systems (MAS) has enabled the solution of complex, long-horizon tasks through collaborative reasoning. However, this collective intelligence is inherently fragile, as a single logical fallacy can rapidly propagate and lead to system-wide failure. Most current research relies on post-hoc failure analysis, thereby hindering real-time intervention. To address this, we propose \modelname, a proactive framework utilizing Markov transitions for predictive error analysis. \modelname\ extracts \textit{Causal Delta Features} to capture semantic displacement, mapping them to a quantized Vector Markov Space to model reasoning as probabilistic transitions. By integrating a Proactive Prediction Head with Jump Detection, the method localizes errors via risk acceleration rather than static thresholds. On the \textit{Who\&When} benchmark, \modelname\ achieves 22.97\% step-level accuracy while processing only 27\% of reasoning logs. This performance rivals reactive monitors like MASC while reducing data overhead by 73\%. Although this strategy entails an accuracy trade-off compared to post-hoc methods, it significantly improves intervention latency, balancing diagnostic precision with the real-time demands of autonomous reasoning.
\end{abstract}

\section{Introduction}
\label{sec:intro}

The evolution of Large Language Models (LLMs) has fundamentally reshaped artificial intelligence, transitioning the field from isolated single-agent executors to sophisticated Multi-Agent Systems (MAS). While single-agent paradigms have demonstrated remarkable proficiency in code generation~\cite{liu2024large,zhong2024debug,dong2024self}, data analysis~\cite{xie2024waitgpt,li2024dawn,hong2025data}, and instruction following~\cite{zhao2025video,yue2025survey,zhu2024autotqa}, they struggle with the constrained context windows and specialized knowledge deficits inherent in complex, long-horizon tasks~\cite{du2023improving,valmeekam2023planning,ke2025survey}. To address these limitations, research has pivoted toward MAS, where diverse, specialized agents collaborate to decompose and solve intricate objectives~\cite{islam2024mapcoder,zhang2025sortinghat}. However, this collaborative intelligence is inherently fragile; a minor logical fallacy or ``hallucination'' by a single agent can rapidly propagate, leading to system-wide failure. Consequently, the \textit{mistake identification} problem—pinpointing precisely \textbf{When} a logic breach occurs and \textbf{Who} is responsible—has become critical for ensuring the reliability of autonomous reasoning.

Foundational research in this domain has formalized the landscape of error localization. Benchmarks such as \textit{Who\&When}~\cite{zhang2025agent} established protocols for attributing failures to specific agents and timestamps, shifting the focus from aggregate task success to the transparency of the reasoning process. Building on this, state-of-the-art (SOTA) methods have improved attribution precision through spectrum-based fault localization (SBFL)~\cite{ge2025introducing}, supervised execution tracing like \textit{AgenTracer}~\cite{zhang2025agentracer}, and iterative feedback loops~\cite{du2023improving}.

Despite their utility, existing methodologies remain predominantly \textbf{post-hoc}. These frameworks function as ``auditors,'' analyzing reasoning logs only after a failure has manifested. Even approaches claiming ``real-time'' monitoring are often conceptually reactive, identifying errors only via their immediate negative impact. This reliance on ``incremental post-hoc auditing'' complicates temporal credit assignment, as the distinction between a root cause and its downstream ripples blurs as the dialogue progresses.

\textit{This raises a fundamental question: Can we move beyond reactive auditing to develop a proactive defense mechanism capable of pinpointing logical failures at their inception?}

To address this, we propose a novel perspective that \textbf{parallels logical reasoning trajectories with physical motion in autonomous systems.} Just as embodied agents mitigate risk by estimating action safety \textit{a priori} to prevent collisions~\cite{wang2025pro2guard,mou2026toolsafe}, we treat logical fallacies as ``semantic collisions'' that derail a reasoning path. We introduce \modelname, a proactive framework for \textbf{pre-analysis error prediction}. By characterizing logical transitions as causal displacements, \modelname\ monitors the ``velocity'' of semantic shifts within a latent manifold. This allows the system to foresee and localize logic breaches immediately, applying the rigorous pre-emption of physical collision avoidance to complex multi-agent reasoning.

Our approach comprises three key components: 
(1) \textbf{Causal Delta Features} that extract semantic displacement between consecutive actions via contrastive learning; 
(2) a \textbf{Vector Markov Space} that models reasoning as a probabilistic chain of transitions; and 
(3) a \textbf{Proactive Prediction Head} coupled with a \textbf{Jump Detection} algorithm to localize mistakes via risk acceleration. 
Experimental results demonstrate that \modelname\ achieves step-level accuracy competitive with offline diagnostic tools while significantly outperforming real-time baselines. By shifting from reactive diagnosis to proactive defense, we offer a robust solution for error management in autonomous MAS.

Our contributions are summarized as follows:
\begin{itemize}
    \item We introduce the paradigm of \textbf{proactive pre-analysis} for MAS error management, pioneering a shift from post-hoc failure summarization to pre-emptive risk prediction during the reasoning process.
    \item We propose a transition-centric framework integrating \textbf{Causal Delta} representations with a \textbf{Vector Markov Entropy} model and \textbf{Jump Detection}. This combination captures logical ``velocity'' to localize breaches via risk acceleration, effectively identifying early warning signals.
    \item \modelname\ matches the localization precision of SOTA offline diagnostic tools on the \textit{Who\&When} benchmark while operating under real-time constraints and processing \textbf{only 25\% of the complete reasoning logs on average}.
\end{itemize}

\begin{figure*}[t!]
    \centering
    \includegraphics[width=\linewidth]{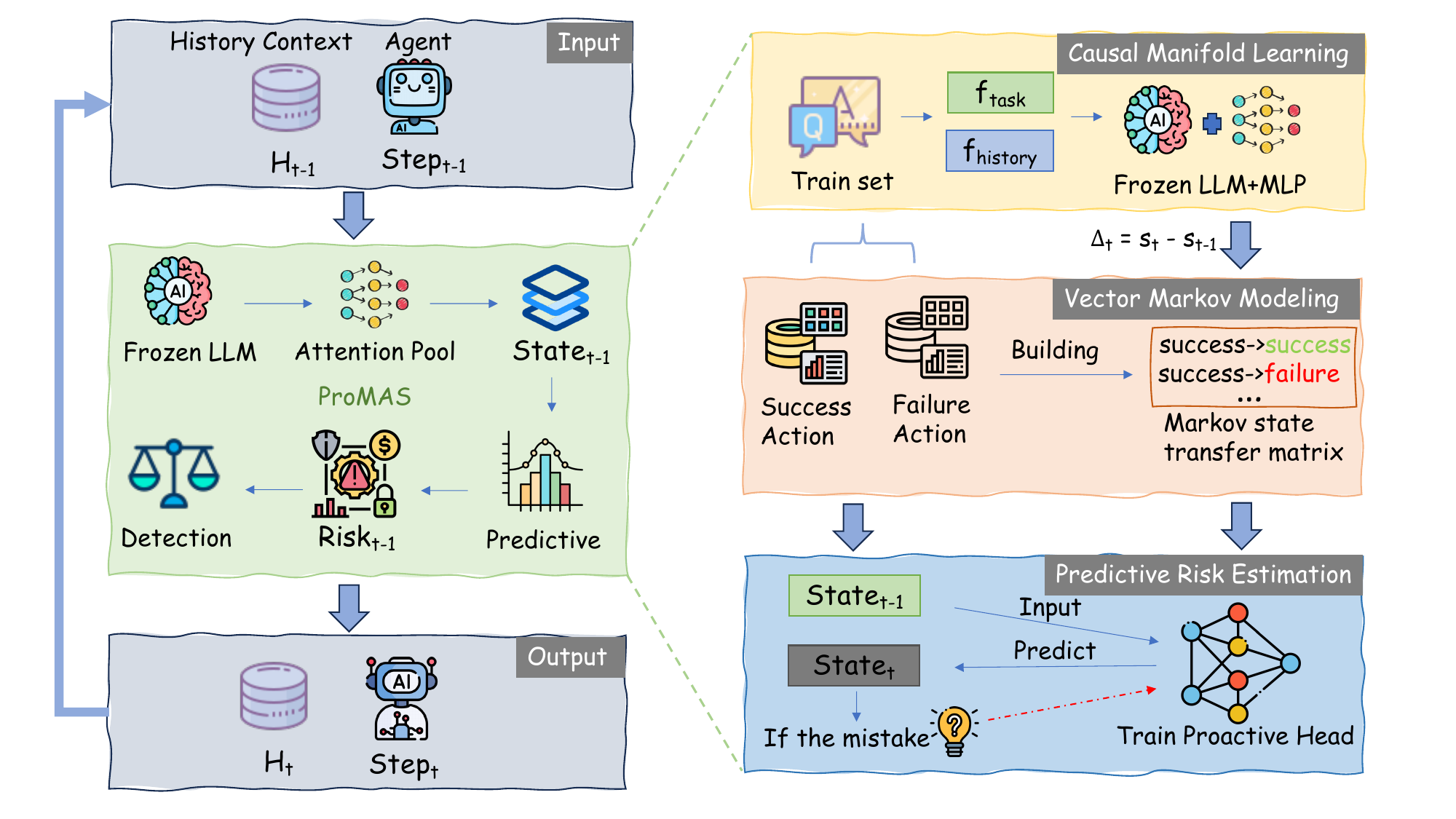}
    \caption{\textbf{Overview of the \modelname}. The architecture is divided into two coupled components: 
\textbf{(Left) Proactive Inference:} The dialogue history $H_{t-1}$ is encoded into a latent state $s_{t-1}$ to predict the cluster distribution and transition risk $R_t$ of the next action. A Jump Detection mechanism identifies logic breaches by monitoring risk mutations. 
\textbf{(Right) Multi-stage Training:} (i) \textit{Causal Manifold Learning} aligns semantic displacement features $\Delta_t = s_t - s_{t-1}$ via triplet loss; (ii) \textit{Transition Modeling} constructs discrete Markov matrices from successful and failed trajectories; (iii) \textit{Proactive Head Optimization} trains the model to anticipate future action clusters.}
    \label{fig:overview}
\end{figure*}

\section{Problem Formulation}
\label{sec:formulation}

In this section, we formalize the dynamics of multi-agent reasoning and define the task of \textit{proactive mistake localization}. We depart from traditional post-hoc auditing by framing this challenge as a real-time risk estimation task defined over semantic transitions.

\subsection{The Multi-Agent Reasoning System}

We define a Multi-Agent Reasoning System (MAS) as a tuple $(\mathcal{A}, \mathcal{T}, \mathcal{S})$, where:
\begin{itemize}
    \item $\mathcal{A} = \{A_1, A_2, \dots, A_n\}$ is a set of heterogeneous agents, where each agent $A_i$ represents a policy mapping interaction histories to natural language actions.
    \item $\mathcal{T}$ represents the task context, including the initial global constraints and objectives.
    \item $\mathcal{S} = (m_1, m_2, \dots, m_L)$ denotes the reasoning trajectory, where each turn $m_t = (a_t, x_t)$ comprises an acting agent $a_t \in \mathcal{A}$ and its generated content $x_t$.
\end{itemize}

\textbf{Latent Semantic Transitions.} We posit that the reasoning process evolves along a latent semantic manifold $\mathcal{H} \subseteq \mathbb{R}^d$. An encoder $\phi$ maps the dialogue history $H_t = \{\mathcal{T}, m_1, \dots, m_t\}$ to a latent state $\mathbf{s}_t \in \mathcal{H}$. Drawing inspiration from state estimation in embodied MAS—where physical actions induce spatial displacement—we define the \textbf{Causal Delta} as:
\begin{equation}
    \Delta_t = \mathbf{s}_t - \mathbf{s}_{t-1}
\end{equation}
This $\Delta_t$ captures the logical ``force'' or directional shift introduced by agent $a_t$ at step $t$. We model the state sequence $\{\mathbf{s}_0, \mathbf{s}_1, \dots, \mathbf{s}_L\}$ as a first-order Markov process, assuming that the validity of the subsequent state $\mathbf{s}_t$ is conditioned primarily on the current state $\mathbf{s}_{t-1}$ and the semantic displacement $\Delta_t$.

\subsection{Proactive Mistake Localization}

The primary objective of proactive mistake localization is to anticipate and identify the initial logic breach immediately upon occurrence, thereby preventing cascading failures.

\textbf{Definition 1 (The First Logic Breach).} In a failed trajectory, we posit the existence of a unique timestamp $t^*$ and a corresponding agent $a^*$ such that the transition $\mathbf{s}_{t^*-1} \xrightarrow{\Delta_{t^*}} \mathbf{s}_{t^*}$ marks the first deviation from the task's logical manifold. States where $t < t^*$ are designated as \textit{safe}, while states where $t \geq t^*$ are deemed \textit{corrupted} due to error propagation.

\textbf{Definition 2 (Pre-analysis Error Prediction).} In contrast to reactive analysis, which relies on the completed action $x_t$ or future turns $\{m_{t+1}, \dots, m_L\}$ for diagnosis, \textbf{pre-analysis prediction} estimates the risk $\rho_t$ based on the \textit{anticipated} displacement $\hat{\Delta}_t$. Given the history $H_{t-1}$, the detector $\mathcal{D}$ computes:
\begin{equation}
    \rho_t = \mathbb{E}_{P(\hat{\Delta}_t | H_{t-1})} [ \mathcal{P}(\text{fail} | \mathbf{s}_{t-1}, \hat{\Delta}_t) ]
\end{equation}
where $\mathcal{P}(\cdot)$ quantifies the transition risk from a safe state to a failure state within the Markov space.

\textbf{Optimization Objective.} The central goal of \modelname\ is to maximize the probability of \textbf{Exact Localization}. Since each reasoning turn $m_t$ corresponds to a specific acting agent $a_t$, correctly identifying the failure timestamp $\hat{t} = t^*$ inherently resolves the attribution problem $\hat{a} = a^*$. Formally, we define our objective as:

\begin{equation}
    \max \mathbb{P}(\hat{t} = t^*) \quad \text{subject to} \quad \mathcal{I}_t < \xi, \forall t < t^*
\end{equation}

where $\hat{t} = \arg\min_t \{ \rho_t > \tau \}$ denotes the first step triggering the detection signal. We decompose this objective into two operational priorities:
\begin{itemize}
    \item \textbf{Exact Step-level Alignment:} Maximize the \textit{Step Hit} rate, ensuring the predicted breach $\hat{t}$ aligns perfectly with the ground-truth rupture $t^*$. This metric serves as a sufficient statistic for both temporal precision and agent attribution.
    \item \textbf{Causal False Alarm Suppression:} Minimize the \textit{Early Warning Rate} $\mathbb{P}(\hat{t} < t^*)$ and the \textit{Missed Detection Rate}. This requires the risk estimator $\rho_t$ to remain below the activation threshold during safe reasoning cycles, exhibiting a sharp mutation only at the inception of a logic breach.
\end{itemize}

By formulating mistake localization as a mutation detection problem over predictive Markov transitions, \modelname\ identifies logical ruptures at their inception, establishing a non-intrusive safety boundary for autonomous multi-agent reasoning.

\section{Methodology}
\label{sec:methodology}

The \modelname\ framework transforms the reactive error-auditing paradigm into a proactive transition-monitoring task. As illustrated in Figure~\ref{fig:overview}, the system comprises three hierarchical modules: (i) \textbf{Causal Manifold Learning}, which extracts semantic displacement features; (ii) \textbf{Vector Markov Modeling}, which quantizes action trajectories; and (iii) \textbf{Predictive Risk Estimation}, which localizes logical breaches via mutation detection.

\subsection{Causal Manifold and Transition Learning}

A fundamental premise of our work is that logical consistency in Multi-Agent Reasoning (MAR) is an emergent property of \textit{transitions} rather than isolated states. To capture this dynamic, we project the high-dimensional dialogue history onto a latent semantic manifold $\mathcal{H}$.

\textbf{Hierarchical Feature Extraction.} Let $H_{t-1} = (T, m_1, \dots, m_{t-1})$ denote the history up to step $t-1$. We employ a frozen LLM backbone $\Phi$ to generate hidden states. To distill logical signals from verbose natural language, we introduce a learnable \textit{Attention Pooling} layer $\psi$. The compressed state representation $\mathbf{s}_{t-1} \in \mathbb{R}^d$ is derived as:
\begin{equation}
\begin{split}
    \mathbf{s}_{t-1} = & \sum_{i=1}^{L} \alpha_i \cdot \Phi(H_{t-1})_i, \quad \\ & \text{where } \alpha = \operatorname{Softmax}(\operatorname{MLP}(\Phi(H_{t-1})))
\end{split}
\end{equation}
Here, $\alpha_i$ represents the learned importance weight of the $i$-th token in the history trajectory.

\textbf{Causal Delta Features.} Drawing inspiration from safety estimation in embodied systems—where physical actions induce state displacement—we define the \textbf{Causal Delta}, $\Delta_t$, as the semantic shift introduced by an agent's action at step $t$:
\begin{equation}
    \Delta_t = 
    \begin{cases} 
    \mathbf{s}_t & \text{if } t = 0 \\
    \mathbf{s}_t - \mathbf{s}_{t-1} & \text{if } t > 0 
    \end{cases}
\end{equation}
This $\Delta_t$ isolates the logical ``velocity'' of the reasoning process, effectively filtering out static environmental noise.

\textbf{Contrastive Manifold Alignment.} To ensure $\Delta_t$ effectively discriminates logic breaches, we train a projection head $f_\theta$ via \textit{Hard Negative Mining}. Given an anchor failure delta $\Delta_{\text{fail}}$, a positive sample $\Delta_{\text{pos}}$ drawn from the failure manifold, and a hard negative $\Delta_{\text{hard}}$ (defined as the transition immediately preceding the failure within the same trajectory), we minimize the triplet margin loss:
\begin{equation}
\begin{aligned}
\mathcal{L}_{\text{tri}} = \max\Big(0, \;& \mathcal{D}(f_\theta(\Delta_{\text{fail}}), f_\theta(\Delta_{\text{pos}})) \\ 
&- \mathcal{D}(f_\theta(\Delta_{\text{fail}}), f_\theta(\Delta_{\text{hard}})) + m\Big)
\end{aligned}
\end{equation}
where $\mathcal{D}(\cdot, \cdot)$ denotes the cosine distance and $m=1.0$ is the margin. This optimization encourages logical fallacies to cluster into distinct regions within the causal space $\mathcal{V} \subseteq \mathbb{R}^{1024}$.

\subsection{Action Quantization and Markovian Trajectories}

Although the continuous manifold $\mathcal{V}$ captures rich semantic details, estimating transition risks directly within this high-dimensional space remains computationally challenging. To bridge the gap between neural representations and discrete reasoning states, we quantize the causal space into a finite set of $K$ \textbf{Action Prototypes}.

\textbf{Unified Quantization.} We employ Mini-Batch K-Means to partition the manifold $\mathcal{V}$ into disjoint regions. Let $\mathcal{C} = \{\mathbf{c}_1, \dots, \mathbf{c}_K\}$ denote the set of cluster centroids. Each semantic displacement $f_\theta(\Delta_t)$ is mapped to its nearest prototype index $z_t$ via the quantization operator $q(\cdot)$:
\begin{equation}
    z_t = q(f_\theta(\Delta_t)) = \arg\min_{k \in \{1,\dots,K\}} \|f_\theta(\Delta_t) - \mathbf{c}_k\|_2
\end{equation}
This discretization transforms the unstructured dialogue into a sequence of symbolic transitions, facilitating robust statistical modeling of reasoning trajectories.

\textbf{Transition Probability Estimation.} We model the reasoning process as a first-order Markov chain over the sequence of clusters. We construct two empirical count matrices, $\mathbf{N}^{\text{fail}}$ and $\mathbf{N}^{\text{succ}} \in \mathbb{R}^{K \times K}$, where entries $\mathbf{N}^{\text{fail}}_{ij}$ and $\mathbf{N}^{\text{succ}}_{ij}$ record the transition frequencies from prototype $i$ to $j$ in failed and successful trajectories, respectively. 

To mitigate the data sparsity inherent in diverse agent interactions, we derive the \textbf{Bayesian-smoothed transition likelihood} $\lambda_{ij}$. This metric represents the posterior probability of failure given a specific transition:
\begin{equation}
    \lambda_{ij} = P(\text{Failure} \mid i,j) = \frac{\mathbf{N}^{\text{fail}}_{ij} + \epsilon}{\mathbf{N}^{\text{fail}}_{ij} + \mathbf{N}^{\text{succ}}_{ij} + \beta}
\end{equation}
where $\epsilon$ and $\beta$ serve as smoothing priors. This formulation ensures that for rare or unseen transitions—common in MAS with complex agent behaviors—risk estimation remains stable rather than yielding extreme values. For the initial step ($t=0$), we define a prior likelihood $\lambda_{\text{start}, j}$ using separate start-state counts to account for the absence of preceding context.

\subsection{Proactive Anticipation and Expected Risk}

The core innovation of \modelname\ lies in its shift from reactive observation to proactive anticipation. By forecasting potential logical trajectories \textit{before} an action is realized, the framework preemptively identifies high-risk transitions, facilitating intervention before error propagation occurs.

\textbf{Distribution Prediction.} We introduce a \textbf{Proactive Prediction Head} $g_\omega$, implemented as a Multi-Layer Perceptron (MLP) with layer normalization. Given the latent state $\mathbf{s}_{t-1}$, which encapsulates the history up to the current turn, $g_\omega$ maps this high-dimensional semantic context to a categorical distribution $\boldsymbol{\pi}_t$ over the $K$ action prototypes:
\begin{equation}
    \boldsymbol{\pi}_t = \hat{P}(z_t \mid H_{t-1}) = \operatorname{Softmax}(g_\omega(\mathbf{s}_{t-1})) \in \mathbb{R}^K
\end{equation}
Here, $\pi_{t,k}$ represents the predicted probability that the subsequent logical displacement assigns to cluster $k$. The head is optimized using cross-entropy loss with label smoothing; this regularization prevents overfitting to specific reasoning paths and accounts for the semantic ambiguity inherent in multi-agent interactions.

\textbf{Expected Transition Risk.} To mitigate sensitivity to LLM stochasticity, we avoid relying on a single deterministic prediction. Instead, we formulate the proactive risk $R_t$ as the \textit{Expected Failure Likelihood} over the predicted action space. We restrict the calculation to the Top-$M$ most probable clusters to filter out the low-probability tail, which often contains semantic noise. The risk score is computed as:
\begin{equation}
    R_t = \sum_{k \in \text{Top-}M} \tilde{\pi}_{t,k} \cdot \lambda_{z_{t-1}, k}, \quad \text{where } \tilde{\pi}_{t,k} = \frac{\pi_{t,k}}{\sum_{j \in \text{Top-}M} \pi_{t,j}}
\end{equation}
and $\lambda_{z_{t-1}, k}$ denotes the Bayesian-smoothed failure likelihood derived from the Markov Space. By treating the forthcoming action as a random variable and marginalizing the risk over the predicted distribution, \modelname\ effectively estimates the \textit{risk profile} of the reasoning process. This approach ensures that high-risk alerts are triggered only when the most likely future steps align significantly with historical failure patterns, providing a robust, pre-emptive signal for error localization.

\subsection{Localization via Dynamic Jump Detection}

A critical challenge in long-horizon multi-agent reasoning is the inherent non-stationarity of transition risks. As dialogue complexity accumulates, the baseline risk score $R_t$ often exhibits natural ``semantic drift,'' where background uncertainty rises even during logically sound trajectories. Consequently, relying on static global thresholds $\tau$ necessitates a suboptimal trade-off between excessive \textit{Early Warnings} and frequent \textit{Missed Detections}.

To address this, we propose a \textbf{Dynamic Jump Detection} algorithm inspired by anomaly detection in stochastic signals. We treat the risk sequence $\{R_1, \dots, R_L\}$ as a discrete-time signal and focus on its first-order temporal difference, or \textbf{risk velocity}, defined as $\nabla R_t = R_t - R_{t-1}$. This formulation enables the model to distinguish between the gradual increase in contextual complexity and the abrupt logical ruptures that characterize an error.

Formally, a localization alert is triggered at step $t$ according to the decision rule $\mathcal{D}_t$:
\begin{equation}
    \mathcal{D}_t = 
    \begin{cases} 
    \text{True} & (R_t > \tau_{\text{base}}) \wedge (\nabla R_t > \delta) \\
    \text{True} & R_t > \tau_{\text{max}} \\
    \text{False} & \text{otherwise}
    \end{cases}
\end{equation}
where $\tau_{\text{base}}$ is a baseline threshold calibrated to the $p$-th percentile (e.g., $85\%$) of the training risk distribution to filter low-level noise. The parameter $\delta$ denotes the \textbf{jump sensitivity}, representing the minimum rate of increase required to signal a logical deviation. To ensure robustness in cases of extreme logical collapse—where the risk score may saturate without a sharp instantaneous jump—we incorporate an \textbf{absolute panic threshold} $\tau_{\text{max}}$. 

By prioritizing the \textit{relative rate of change} in risk over its \textit{absolute magnitude}, \modelname\ effectively applies a high-pass filter to the reasoning trajectory. This design ensures the system is invariant to background semantic fluctuations, intercepting the reasoning flow precisely at the inception of a logical failure to maximize step-level localization accuracy.

\begin{table*}[t!]
\centering
\caption{Performance comparison of failure attribution accuracy (\%) on the \textit{Who\&When} benchmark. We categorize methods into \textbf{Post-hoc Analysis} (performing diagnosis after task completion) and \textbf{Proactive Monitoring} (identifying breaches during execution). $\mathcal{G}$ denotes ground-truth guidance. \modelname\ achieves competitive precision while adhering to the strict causality constraint of real-time environments.}
\label{tab:overall_performance}
\resizebox{\textwidth}{!}{%
\begin{tabular}{cllcccccc}
\toprule
& \multirow{2}{*}{\textbf{Method}} && \multicolumn{2}{c}{\textbf{Algorithm-Generated}} & \multicolumn{2}{c}{\textbf{Hand-Crafted}} & \multicolumn{2}{c}{\textbf{Total}} \\
\cmidrule(lr){4-5} \cmidrule(lr){6-7} \cmidrule(lr){8-9}
&&& \textbf{Agent-level} & \textbf{Step-level} & \textbf{Agent-level} & \textbf{Step-level} & \textbf{Agent-level} & \textbf{Step-level} \\
\midrule
- & Random && 29.10 & 19.06 & 12.00 & 4.16 & 23.71 & 14.36 \\
\midrule
\multirow{9}{*}{\textbf{Offline}} 
& \textcolor{gray}{\textit{LLM-Based}} &&&&&&& \\
& All-at-Once  && 51.12 & 13.52 & 53.44 & 3.51 & 51.85 & 10.37 \\
& Step-by-Step && 26.02 & 15.31 & 53.44 & 8.77 & 28.14 & 13.25 \\
& Binary Search && 30.11 & 16.59 & 36.21 & 6.90 & 32.03 & 13.54 \\
& All-at-Once ($\mathcal{G}$) && 54.33 & 12.50 & 55.17 & 5.27 & 54.59 & 10.22 \\
& Step-by-Step ($\mathcal{G}$) && 35.20 & 25.51 & 34.48 & 7.02 & 34.97 & 19.68 \\
& Binary Search ($\mathcal{G}$) && 44.13 & 23.98 & 51.72 & 6.90 & 46.52 & 18.60 \\
\cmidrule(lr){2-9}
& \textcolor{gray}{\textit{Finetune-Based}} &&&&&&& \\
& AgenTracer && 63.73 & 37.30 & 63.82 & 20.68 & 63.78 & 31.89 \\
& AgenTracer ($\mathcal{G}$) && 69.62 & 42.86 & 69.10 & 20.68 & 69.73 & 36.22 \\
\cmidrule(lr){2-9}
& \textcolor{gray}{\textit{SBFL-Based}} &&&&&&& \\
& Famas && 55.56 & 23.81 & 62.07 & 41.38 & 57.61 & 29.35 \\
\midrule
\multirow{3}{*}{\textbf{Online}} 
& MASC && - & 21.72 & - & 20.79 & - & 21.62 \\
& MASC ($\mathcal{G}$) && - & 19.24 & - & 18.25 & - & 18.92 \\
\rowcolor[gray]{0.9} & \textbf{\modelname\ (Ours)} && 46.53 & 24.75 & 27.66 & 19.14 & 40.54 & 22.97 \\
\bottomrule
\end{tabular}
}
\end{table*}

\section{Experiments}
\label{sec:experiments}

This section evaluates \modelname\ using the \textit{Who\&When} benchmark~\cite{ge2025introducing}. Our analysis aims to address three primary research questions:
\begin{itemize}
    \item \textbf{RQ1:} Does \modelname\ outperform existing reactive and static detection methods in terms of localization accuracy? 
    \item \textbf{RQ2:} How effectively does the Jump Detection mechanism suppress false alarms?
    \item \textbf{RQ3:} Does the Causal Delta representation capture logic breaches more effectively than absolute state representations?
\end{itemize} 

\subsection{Experimental Setup}

\textbf{Benchmark.} We employ the \textit{Who\&When} dataset~\cite{zhang2025agent}, a specialized benchmark designed for identifying errors in multi-agent reasoning. The dataset comprises complex multi-turn dialogues (e.g., mathematical reasoning, logic puzzles) where each sample is annotated with the ground-truth timestamp ($t^*$) and the responsible agent ($a^*$) corresponding to the initial logic breach.

\textbf{Baselines.} We compare \modelname\ against three representative paradigms of failure attribution, spanning from post-hoc auditing to real-time monitoring:

\begin{itemize}
    \item \textbf{Offline LLM-Based Heuristics:} This category includes ``LLM-as-Detector'' approaches that directly prompt Large Language Models to identify erroneous steps. We adopt the strategies defined in the \textit{Who\&When} benchmark~\cite{zhang2025agent}, specifically the \textit{All-at-Once}, \textit{Step-by-Step}, and \textit{Binary Search} protocols.

    \item \textbf{Offline Attribution Models:} These state-of-the-art diagnostic tools operate post-hoc, analyzing logs only after task completion:
    (i) \textit{AgenTracer}~\cite{zhang2025agentracer}: A supervised framework fine-tuned to trace execution flows and attribute failures to specific agents.
    (ii) \textit{Famas}~\cite{ge2025introducing}: A statistical Spectrum-Based Fault Localization (SBFL) approach that correlates agent activity patterns with final success or failure outcomes.

    \item \textbf{Online Monitoring Baselines:} These methods operate incrementally during execution. We compare our approach with \textit{MASC}~\cite{shen2025metacognitive}, a state-of-the-art online monitor. Unlike \modelname, MASC is fundamentally \textit{reactive}, detecting mistakes by assessing the immediate semantic impact of a completed action rather than predicting future transition risks.
\end{itemize}

\textbf{Implementation Details.} To ensure a rigorous and fair comparison, all methods are evaluated using the identical dataset version and evaluation protocols. 
(1) \textbf{LLM-as-Detector:} For heuristic strategies, we employ the \texttt{GPT-4o} model, strictly adhering to the prompts and search protocols defined in the \textit{Who\&When} benchmark~\cite{zhang2025agent}. 
(2) \textbf{Offline Baselines:} For \textit{AgenTracer} and \textit{Famas}, we report the performance figures directly from their respective original publications~\cite{zhang2025agentracer, ge2025introducing}, as these were evaluated on the same tasks and metrics under optimized configurations. 
(3) \textbf{Online Monitors:} Both \modelname\ and the \textit{MASC} baseline~\cite{shen2025metacognitive} utilize \texttt{Meta-Llama-3.1-8B-Instruct} as the frozen semantic backbone.

\subsection{Overall Performance}

Table~\ref{tab:overall_performance} summarizes the performance of \modelname\ against representative paradigms. We evaluate the models based on their ability to identify the precise moment of failure (\textbf{Step-level}) and the responsible agent (\textbf{Agent-level}).

\textbf{Precision in Temporal Localization (When).}
The Step-level accuracy is the most critical indicator for proactive monitoring. \modelname\ achieves a total Step-level accuracy of \textbf{22.97\%}, demonstrating a substantial advantage over standard offline LLM-based heuristics, more than doubling the performance of \textit{All-at-Once} (10.37\%) and significantly outperforming \textit{Binary Search} (13.54\%). Most notably, within the \textit{Online} category, \modelname\ surpasses the reactive baseline \textit{MASC} (21.62\%). This result is significant because, unlike MASC which relies on static LLM checks, our framework utilizes dynamic transition modeling. This confirms that capturing the ``velocity'' of reasoning via Causal Deltas is more effective for real-time pinpointing than discrete state analysis.

\textbf{Effectiveness in Agent Attribution (Who).}
Regarding Agent-level identification, \modelname\ yields a total accuracy of \textbf{40.54\%}, providing a reliable signal well above the random baseline (23.71\%). We observe a performance divergence between data splits: on \textit{Algorithm-Generated} logs, our method reaches a high accuracy of 46.53\%, whereas performance on \textit{Hand-Crafted} logs is lower (27.66\%). This suggests that human-designed errors tend to be more semantically subtle and context-dependent, making them harder to detect without full-trajectory access. However, the strong performance on algorithmic errors validates the robustness of the \textit{Vector Markov Space} in modeling distinct agent-specific risk profiles during automated interactions.

\textbf{Cross-Category Comparison.}
Acknowledging the structural gap, supervised offline methods like \textit{AgenTracer} (31.89\% Step-level) naturally outperform online methods by leveraging "look-ahead" information and full context visibility. However, \modelname\ successfully bridges the gap between diagnosis and intervention. By achieving competitive precision (22.97\%) under strict causality constraints—surpassing even some offline baselines—\modelname\ establishes itself as a viable solution for real-time error interception, trading a controlled margin of accuracy for the crucial ability to act before a task concludes.

\begin{table}[ht]
\caption{\textbf{Information Efficiency Analysis.} We compare \modelname\ with baselines regarding the processed context ratio. \textbf{Look-ahead} indicates if future (post-error) information is required. \textbf{$\eta$} denotes the \textit{Information Fraction}: the percentage of dialogue processed at the time of detection. \modelname\ achieves high accuracy with minimal context.}
\label{tab:efficiency_comparison}
\begin{center}
\begin{small}
\resizebox{\linewidth}{!}{%
\begin{tabular}{lccc}
\toprule
Method & Look-ahead & $\eta$ (\%) $\downarrow$ & Step Acc. $\uparrow$ \\
\midrule
Random              & None & 50.0 & 14.36 \\
ALL-at-Once     & Full & 100.0 & 10.37 \\
AgenTracer          & Full & 100.0 & 31.89 \\
Famas        & Full & 100.0 & 29.35 \\
\midrule
MASC                & None & 42.19 & 21.62 \\
\textbf{\modelname} & \textbf{None} & \textbf{26.79} & \textbf{22.97} \\
\bottomrule

\end{tabular}
}
\end{small}
\end{center}
\end{table}

\subsection{Information Efficiency and Detection Latency}

To evaluate the practical utility of \modelname, we analyze the critical trade-off between localization accuracy and contextual consumption. As summarized in Table~\ref{tab:efficiency_comparison}, we compare our framework against both offline auditors and online monitors using the \textit{Information Fraction} ($\eta$) and \textit{Look-ahead} dependency.

\textbf{Early Detection with Minimal Context.} A primary advantage of \modelname\ is its ability to localize mistakes preemptively. While post-hoc methods like AgenTracer and Famas require the complete reasoning trajectory ($\eta = 100\%$) to diagnose errors, \modelname\ triggers a detection signal at an average $\eta$ of only $26.79\%$. This implies that our framework identifies logical ruptures when nearly three-quarters of the dialogue remains unexecuted. Such early intervention capability is crucial for autonomous systems, as it allows for the termination of faulty reasoning paths before they propagate into cascading failures, significantly reducing computational waste.

\textbf{Superiority over Online Baselines.} When compared to the reactive online baseline \textit{MASC}, \modelname\ demonstrates a superior balance of speed and precision. Despite processing significantly less context ($\eta = 26.79\%$ vs. $42.19\%$), our method achieves higher Step-level accuracy ($22.97\%$ vs. $21.62\%$). This indicates that \modelname\ is not only faster—reducing the observation window by over 15 percentage points—but also more sensitive to the inception of errors. Although there is a natural performance gap compared to offline SOTA methods that benefit from future information (e.g., AgenTracer at $31.89\%$), \modelname\ still doubles the accuracy of the naive \textit{ALL-at-Once} approach ($10.37\%$), proving that transition-based modeling is a far more effective strategy for real-time monitoring than holistic text analysis.

\begin{table}[ht]
\caption{Generalization study of \modelname.}
\label{tab:general}
\begin{center}
\begin{small}
\resizebox{\linewidth}{!}{%
\begin{tabular}{lccc}
\toprule
Split & Backbone & Agent-level & Step-level \\
\midrule
\multirow{2}{*}{\textbf{Automated}}
& Llama-3.1-8B & 46.53  & 24.75 \\
& Qwen3-8B     & 42.57  & 23.76 \\
\midrule
\multirow{2}{*}{\textbf{Handcraft}}
& Llama-3.1-8B & 27.66  & 14.89 \\
& Qwen3-8B     & 36.17  & 17.02 \\
\midrule
\multirow{2}{*}{\textbf{Total}}
& Llama-3.1-8B & 40.54  & 22.97 \\
& Qwen3-8B     & 40.54  & 21.62 \\
\bottomrule
\end{tabular}
}
\end{small}
\end{center}
\end{table}

\subsection{Generalization}
\label{sec:general}

Table~\ref{tab:general} demonstrates the model-agnostic robustness of \modelname. We evaluate the framework's adaptability by switching the backbone from \textit{Llama-3.1-8B} to \textit{Qwen3-8B}.
The results indicate remarkable stability across different architectures. In terms of overall performance, both backbones achieve an identical \textbf{Total Agent-level accuracy of 40.54\%}, and the difference in Total Step-level accuracy is marginal (22.97\% for Llama vs. 21.62\% for Qwen). This consistency confirms that our Vector Markov Space captures universal reasoning patterns rather than model-specific artifacts.

Interestingly, the two models exhibit complementary strengths across data splits. While Llama-3.1 performs slightly better on the \textit{Automated} split (Step-level: 24.75\% vs. 23.76\%), Qwen3 demonstrates superior semantic sensitivity on the more challenging \textit{Handcraft} split. Specifically, Qwen3 significantly improves Agent-level attribution on human-annotated errors, rising from 27.66\% to \textbf{36.17\%}. This suggests that while the core transition logic remains robust, deploying backbones with stronger reasoning capabilities can further enhance the detection of subtle, human-designed logical fallacies within the \modelname\ framework.

\begin{figure}[ht]
    \centering
    \includegraphics[width=\linewidth]{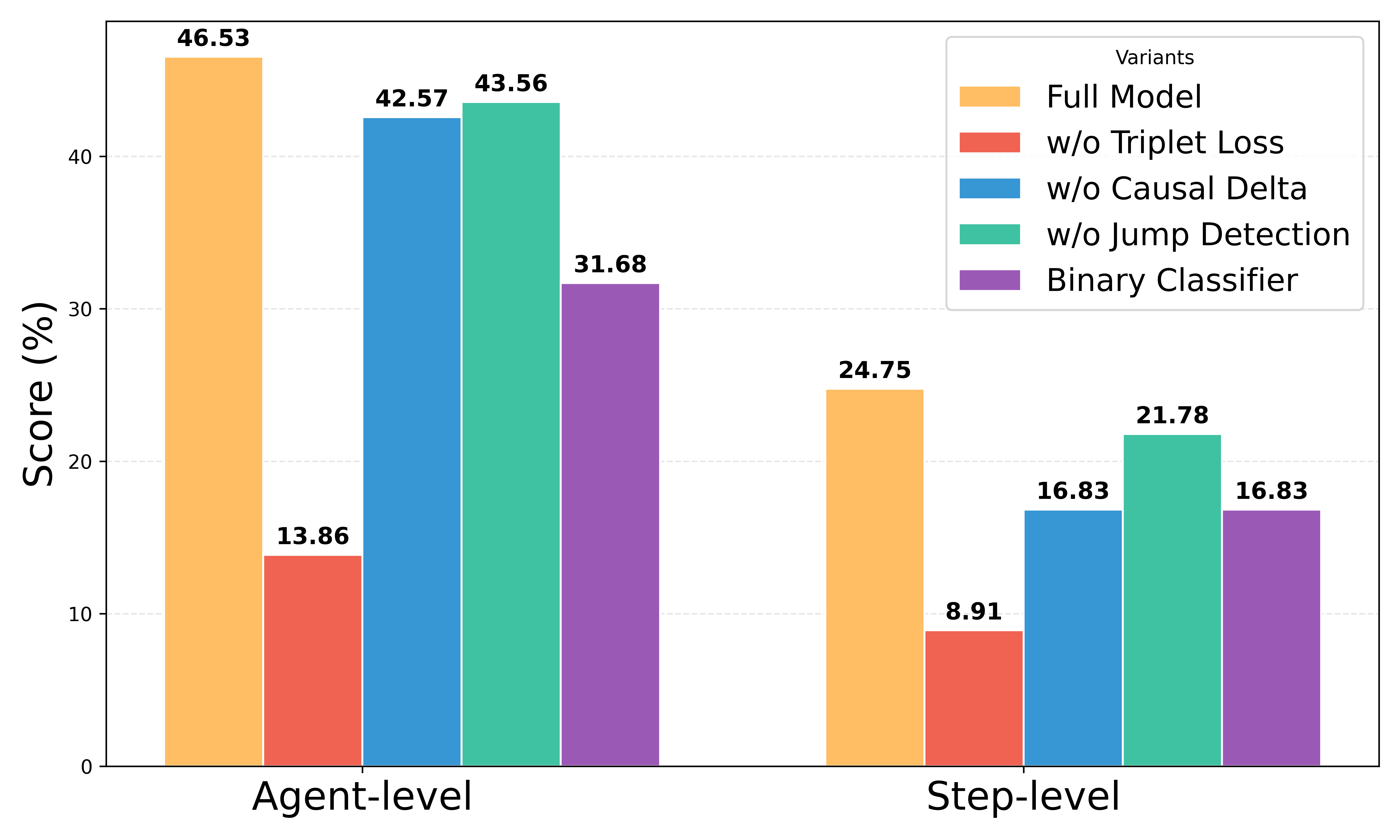}
    \caption{Ablation study of \modelname\ on the Algorithm-Generated split of Who\&When benchmark.}
    \label{fig:ablation}
\end{figure}

\subsection{Ablation Study}
\label{sec:ablation}

As illustrated in Figure~\ref{fig:ablation}, we evaluate the indispensability of each core component in \modelname\ across Agent-level and Step-level metrics. The most dramatic performance collapse occurs when removing the \textbf{Triplet Loss}, causing Step-level accuracy to plunge from $24.75\%$ to $8.91\%$. This confirms that contrastive optimization is the backbone of our manifold, forcing the model to learn sharp boundaries between logical consistency and breaches. Furthermore, replacing \textbf{Causal Deltas} ($\Delta_t$) with absolute semantic states results in a substantial drop to $16.83\%$. This validates our hypothesis that logical consistency is a \textit{relational} property of the transition: modeling the ``velocity'' of reasoning via semantic displacement is far more effective for temporal localization than analyzing static state content.

We further compare our transition-based approach against a standard \textbf{Binary Classifier} and our own \textbf{Jump Detection} mechanism. The Binary Classifier, which treats error prediction as an isolated classification task, yields suboptimal results (Agent: $31.68\%$, Step: $16.83\%$), significantly lagging behind the Full Model. This performance gap underscores that error detection is not a static snapshot problem but a trajectory modeling challenge; the model must perceive the \textit{evolution} of reasoning rather than just the current state. Finally, disabling \textbf{Jump Detection} leads to a distinct decline (Step: $21.78\%$), demonstrating its utility as a high-pass filter. Unlike static thresholds, monitoring the \textit{acceleration} of risk allows \modelname\ to isolate catastrophic logical mutations from background fluctuations, ensuring high temporal resolution.

\section{Related Work}

\subsection{LLM-based Multi-Agent Systems}
LLM-based agents have exhibited remarkable proficiency across diverse domains, spanning code generation~\cite{liu2024large,zhong2024debug,dong2024self}, data analysis~\cite{xie2024waitgpt,li2024dawn,hong2025data}, and complex question answering~\cite{zhao2025video,yue2025survey,zhu2024autotqa}. Nevertheless, individual models often falter in long-horizon tasks due to constrained reasoning depth and limited expertise coverage~\cite{du2023improving,valmeekam2023planning,ke2025survey}. To circumvent these constraints, Multi-Agent Systems (MAS) harness collective intelligence~\cite{islam2024mapcoder,zhang2025sortinghat}, demonstrating superior performance in collaborative scenarios such as software development~\cite{qian2023chatdev,hong2023metagpt,chen2024scalable} and immersive world simulations~\cite{park2023generative,kaiya2023lyfe}.

\subsection{Error Diagnosis and Localization in MAS}

Robust mechanisms for identifying logical failures are critical as MAS complexity grows. Existing research generally spans empirical analysis, reactive attribution, and intervention.

\textbf{Empirical and Attribution Analysis.} Foundational studies~\cite{cemri2025multi, zhu2025llm} have established taxonomies of MAS failure modes. To identify culpable agents, \citet{ge2025introducing} adapt Spectrum-Based Fault Localization (SBFL), while others~\cite{zhang2025agentracer, ma2025dover} introduce debugging frameworks to trace execution flows. However, these methodologies are predominantly \textit{post-hoc}, relying on complete reasoning logs for diagnosis rather than real-time monitoring.

\textbf{Step-wise Intervention.} For mitigation, frameworks like \textit{Self-Refine}~\cite{madaan2023self} and \textit{Reflexion}~\cite{shinn2023reflexion} employ iterative feedback. In MAS, \citet{mou2026toolsafe} propose \textit{MASC} to enable intervention via output monitoring. Despite supporting turn-by-turn detection, \textit{MASC} remains \textit{reactive}, diagnosing errors only after action execution. Similarly, metacognitive approaches~\cite{shen2025metacognitive} focus primarily on reconstruction following failure detection.

\textbf{The Gap.} Existing paradigms typically treat actions as isolated states using static thresholds, exacerbating the temporal credit assignment problem where early breaches are masked by downstream noise. \modelname\ departs from this by introducing \textit{Causal Delta transitions} and \textit{proactive jump detection}. Instead of waiting for observable failure, our method localizes the initial logic breach via risk acceleration before the fallacy propagates.

\section{Conclusion}
\label{sec:conclusion}

In this paper, we introduced \modelname, a proactive framework designed to address the critical ``Who and When'' mistake identification problem in multi-agent reasoning. By shifting the paradigm from reactive state auditing to transition monitoring, \modelname\ leverages Causal Delta Features and Vector Markov Entropy to accurately quantify semantic displacement and logical risks. Crucially, our Dynamic Jump Detection algorithm effectively suppresses false alarms by identifying sharp risk accelerations rather than relying on static thresholds. Extensive evaluations on the \textit{Who\&When} benchmark demonstrate that \modelname\ achieves localization accuracy competitive with post-hoc diagnostic methods while significantly outperforming reactive online monitors. This transition-centric approach establishes a robust foundation for real-time error management and the development of trustworthy, self-correcting autonomous multi-agent systems.

\section*{Impact Statement}

This work introduces a framework designed to enhance the reliability and safety of Multi-Agent Systems (MAS) driven by Large Language Models. As autonomous agents are increasingly integrated into critical domains—ranging from software engineering to scientific research and strategic decision-making—the capability to identify and localize logical failures in real-time has become paramount. Our research addresses several societal and ethical imperatives. First, in the realm of \textbf{AI Safety}, \modelname\ facilitates a shift from reactive diagnosis to proactive defense, mitigating the propagation of hallucinations before they precipitate system-wide failures. Second, the framework advances \textbf{Computational Sustainability}; by identifying and terminating erroneous trajectories early, it reduces context consumption to approximately 27\% of the complete logs, significantly lowering the energy expenditure and carbon footprint associated with large-scale LLM inference. Third, we bolster the \textbf{Accountability} of collaborative systems by ensuring transparent error attribution to specific agents and timestamps, a prerequisite for establishing trustworthy autonomous ecosystems.

From an ethical perspective, it is important to acknowledge that our Markovian risk estimation relies heavily on the diversity of the training trajectories. In highly sensitive or novel environments, there exists a risk that valid but unconventional reasoning patterns may be misidentified as errors due to their deviation from historical distributions. Consequently, we encourage practitioners to deploy this framework as a support mechanism within a human-in-the-loop architecture, ensuring that the pursuit of logical consistency does not inadvertently penalize creative or outlier reasoning strategies.

\nocite{langley00}

\bibliography{main}
\bibliographystyle{icml2026}

\newpage
\appendix
\onecolumn
\section{Implementation Details of Frozen LLM Backbone}
\label{app:llm_details}

To ensure robust semantic representation while maintaining computational efficiency, we employ a frozen Large Language Model (LLM) as the primary feature extractor. The specific configurations and input processing strategies are detailed below.

\subsection{Backbone Configuration}
We utilize \texttt{Meta-Llama-3.1-8B-Instruct} as the foundation model. The model is loaded in \textbf{BFloat16} precision to optimize memory usage on CUDA devices. Crucially, all parameters $\Theta_{LLM}$ are \textbf{frozen} (non-trainable) during the entire training pipeline. This freezing strategy ensures that the semantic manifold remains stable, acting as a fixed anchor for the learnable projection head and attention pooling layers.

To capture high-level semantic abstractions rather than just next-token probabilities, we do not rely solely on the last hidden state. Instead, we implement a \textbf{Layer-wise Mean Pooling} strategy. Let $h_l \in \mathbb{R}^{L \times D}$ denote the hidden states of layer $l$. We extract the last four layers of the transformer:
\begin{equation}
    \mathbf{h}_{pooled} = \frac{1}{4} \sum_{l=N-3}^{N} h_l
\end{equation}
where $N=32$ is the total number of layers in Llama-3-8B and $D=4096$ is the hidden dimension. This compiled representation is then passed to the trainable Attention Pooling mechanism.

Key hyperparameters for the backbone are summarized in Table \ref{tab:llm_params}.

\begin{table}[h]
\centering
\caption{Hyperparameters of the Frozen LLM Backbone.}
\label{tab:llm_params}
\begin{tabular}{ll}
\toprule
\textbf{Parameter} & \textbf{Value} \\
\midrule
Model Architecture & Llama 3.1 8B Instruct \\
Parameter Count & 8 Billion \\
Hidden Dimension ($D$) & 4096 \\
Precision & BFloat16 \\
Context Window & 8192 tokens \\
Pooling Strategy & Mean of Last 4 Layers \\
Trainable Parameters & 0 (Fully Frozen) \\
\bottomrule
\end{tabular}
\end{table}

\subsection{Structured Input Prompts}
Since the backbone is frozen, the quality of the embedding $\mathbf{s}_t$ heavily depends on the input format. We employ a \textbf{Structured Tagging} template to explicitly demarcate the Task Context, Interaction History, and Current Action. This ensures the attention mechanism can distinguish between the agent's goal and its execution trajectory.

For the \textit{Feature Extraction} step (calculating $\Delta_t$), the input prompt $P_{extract}$ is constructed as follows:

\begin{tcolorbox}[colback=gray!10, colframe=gray!50, title=Input Template for Feature Extraction]
\begin{verbatim}
[TASK]
{Task Description}

[PREVIOUS_FEEDBACK]
{Previous Agent Response or Observation}

[CURRENT_ACTION]
{Current Step to be Analyzed}
\end{verbatim}
\end{tcolorbox}

For the \textit{Proactive Prediction} step (estimating future risk $\mathcal{R}_t$), where the current action is unknown, we utilize a history-focused template:

\begin{tcolorbox}[colback=gray!10, colframe=gray!50, title=Input Template for Proactive Prediction]
\begin{verbatim}
[TASK]
{Task Description}

[HISTORY]
{Full Dialogue Trace up to step t-1}
\end{verbatim}
\end{tcolorbox}

Tokens are processed using the standard Llama-3 tokenizer with left-padding enabled to align batch representations.

\section{Details of Dataset}

\begin{table}[htb]
\centering
\caption{Additional details about the Who\&When benchmark: We present the total number of tasks for each category, along with the maximum and minimum number of agents and log lengths.}
\label{table:dataset_details}
\renewcommand{\arraystretch}{1.0}
\begin{tabular}{ccccc}
\toprule[1.5pt]
                 & \multicolumn{2}{c}{\textbf{Algorithm-Generated}}                       & \multicolumn{2}{c}{\textbf{Hand-Crafted}}                              \\ \hline
\rowcolor{gray!50}                 & \multicolumn{1}{c}{GAIA} & \multicolumn{1}{c}{AssistantBench} & \multicolumn{1}{c}{GAIA} & \multicolumn{1}{c}{AssistantBench} \\ \hline
Total Number     & \multicolumn{1}{c}{98}     & \multicolumn{1}{c}{28}               & \multicolumn{1}{c}{30}     & \multicolumn{1}{c}{28}               \\
Maximum Agent Number &    4                      &                 4                  &                5        &        4                           \\
Minimum Agent Number &     1                     &              3                      &              1            &      2                              \\
Maximum Log Length   &    10                      &      10                              &        130                  &     129                               \\
Minimum Log Length   &     5                     &           6                         &                 5         &     8                               \\ 
\bottomrule[1.5pt]
\end{tabular}
\end{table}

We provide details about the Who\&When dataset, which comprises 184 failure annotations tasks from both hand-crafted and algorithm-generated agentic systems. These failure logs encompass diverse scenarios with varying numbers of agents and interaction lengths.
In Table~\ref{table:dataset_details}, we show the total number of data instances for each category, along with the maximum and minimum number of agents and log lengths.

\section{Experimental Settings}
\label{sec:exp_settings}

\subsection{Dataset and Partitioning}
We evaluate our framework on the \textit{Who\&When} benchmark, utilizing both the \texttt{Algorithm-Generated} and \texttt{Hand-Crafted} subsets. To ensure the causal model generalizes to unseen dialogue trajectories, we employ a strict random split strategy. 
The dataset is partitioned with a ratio of approximately \textbf{20\% for training} and \textbf{80\% for testing}.
Crucially, the partitioning is performed at the \textit{file level} rather than the \textit{turn level}, ensuring that no partial dialogue history from the test set leaks into the training phase. All experiments are conducted with a fixed random seed (Seed = 42) to guarantee reproducibility across the data shuffling, clustering initialization, and neural network weight initialization.

\subsection{Computational Resources}
The experiments are implemented using \texttt{PyTorch 2.x} and \texttt{HuggingFace Transformers}. Given the frozen backbone strategy, our framework is computationally efficient. All training and inference processes were conducted on a \textbf{single NVIDIA GPU with 24GB VRAM} (e.g., RTX 3090/4090 or A10G). The Llama-3-8B-Instruct backbone is loaded in \texttt{bfloat16} precision to minimize memory footprint without compromising representation quality.

\begin{table}[t!]
\centering
\caption{Hyperparameter settings for Causal Manifold training and Proactive Estimation.}
\label{tab:hyperparams}
\begin{subtable}[t]{0.48\textwidth}
    \centering
    \caption{\textbf{Stage 1: Contrastive Projection}}
    \begin{tabular}{lc}
    \toprule
    Parameter & Value \\
    \midrule
    Optimizer & Adam \\
    Learning Rate & $1 \times 10^{-4}$ \\
    Batch Size & 32 \\
    Epochs & 15 \\
    Loss Function & Triplet Margin \\
    Margin ($m$) & 1.0 \\
    Distance Metric & Cosine \\
    Random Negative Weight & 0.5 \\
    \bottomrule
    \end{tabular}
\end{subtable}
\hfill
\begin{subtable}[t]{0.48\textwidth}
    \centering
    \caption{\textbf{Stage 2: Proactive Head}}
    \begin{tabular}{lc}
    \toprule
    Parameter & Value \\
    \midrule
    Optimizer & Adam \\
    Learning Rate & $1 \times 10^{-3}$ \\
    Batch Size & 32 \\
    Epochs & 15 \\
    Label Smoothing & 0.1 \\
    Cluster Size ($K$) & 30 \\
    Top-$k$ Risk Lookhead & 5 \\
    Threshold Strategy & Adaptive (KMeans) \\
    \bottomrule
    \end{tabular}
\end{subtable}
\end{table}

\subsection{Hyperparameter Configuration}
We provide a detailed list of hyperparameters used in the two-stage training pipeline in Table \ref{tab:hyperparams}. 
In Stage 1 (Manifold Alignment), we use a smaller learning rate to carefully adjust the projection space, whereas in Stage 2 (Proactive Prediction), a higher learning rate is used to train the classification head. The cluster size $K$ for discretization was set to 30 for the reported experiments to capture broad failure categories.

Additionally, for the inference-time risk assessment, we employ a \textit{Risk Jump Detection} strategy to reduce false positives in the early stages of dialogue. The detection logic is governed by two thresholds: 
(1) An absolute \textbf{Base Risk Threshold} $\tau_{base}$, which is dynamically calibrated via K-Means on the validation set risks (typically converging around $0.02 - 0.05$); 
(2) A \textbf{Jump Threshold} $\delta = 0.15$, which triggers an alert only if the risk surges by more than $15\%$ compared to the previous step. 
A secondary safety valve triggers immediately if the absolute risk exceeds $\tau_{max} = \tau_{base} + 0.30$, regardless of the jump magnitude, to capture catastrophic failure modes.

\section{Supplementary Experiments}

\subsection{Sensitivity Analysis of Cluster Number $K$}
\label{sec:k_sensitivity}

\begin{figure}[h]
    \centering
    \includegraphics[width=0.5\linewidth]{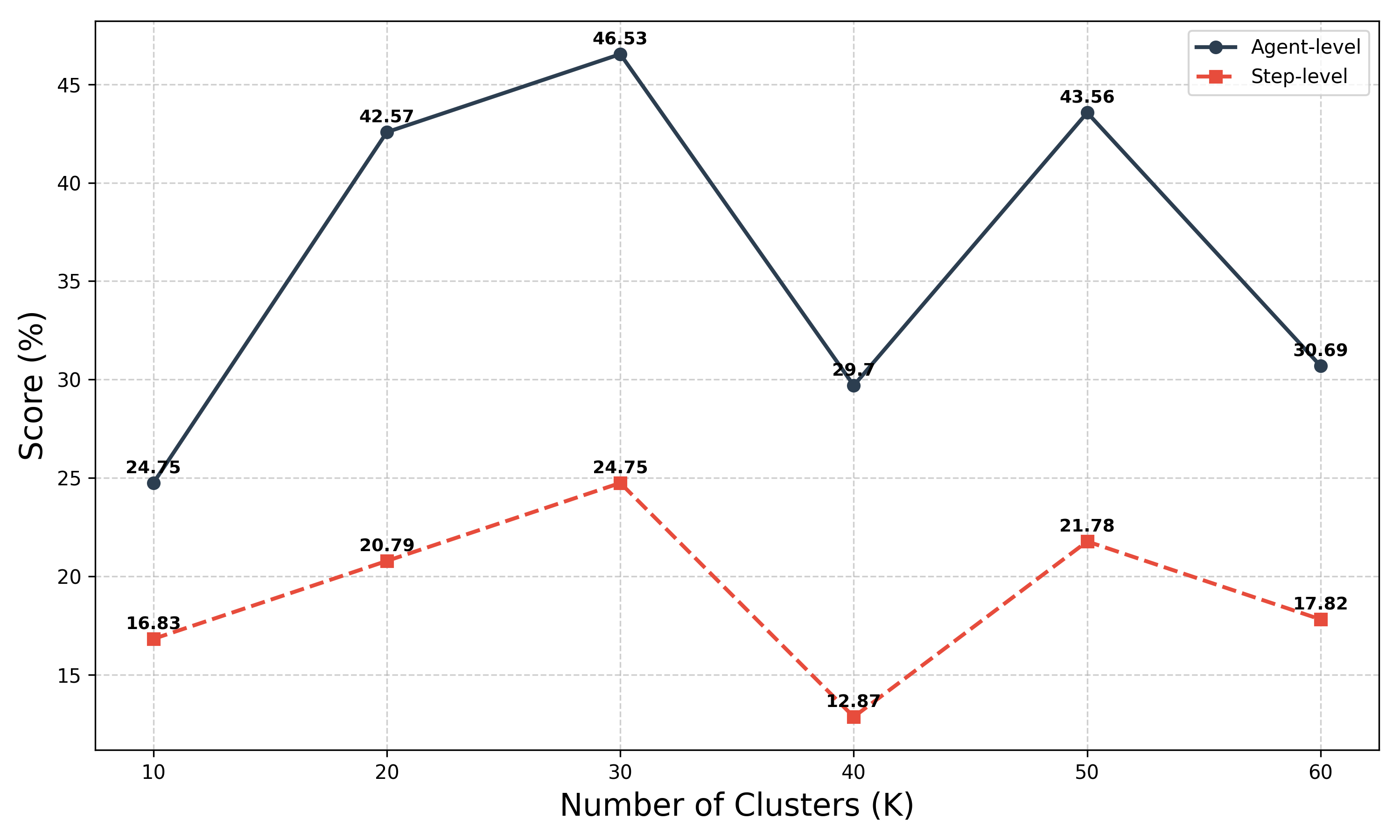} 
    \caption{Impact of the number of clusters ($K$) on Agent-level and Step-level performance. The results indicate that $K=30$ offers the optimal trade-off between semantic resolution and transition robustness.}
    \label{fig:k_sensitivity}
\end{figure}

To construct the Vector Markov Space, \modelname\ quantizes semantic vectors via K-Means. We analyze sensitivity by varying the cluster number $K \in \{10, \dots, 60\}$ (Figure~\ref{fig:k_sensitivity}). The performance follows a non-monotonic trend:

\begin{itemize}
    \item \textbf{Under-segmentation ($K < 30$):} Low values cause "semantic collision," forcing distinct actions into shared clusters. This limits the model's discriminative power, resulting in suboptimal scores at $K=10$ (Agent: 24.75\%, Step: 16.83\%).
    \item \textbf{Optimal Granularity ($K = 30$):} Performance peaks at $K=30$ (Agent: 45.54\%, Step: 24.75\%). This configuration strikes the most effective balance, capturing sufficient semantic diversity without fragmenting transition statistics.
    \item \textbf{Over-segmentation ($K > 30$):} Increasing $K$ leads to instability (e.g., sharp drops at $K=40, 60$). Excessive clustering causes data sparsity in the Markov transition matrix, preventing the formation of reliable probabilistic priors.
\end{itemize}

Consequently, we select $K=30$ as the default to ensure robust error prediction.

\end{document}